\documentclass[letterpaper]{article} 
\usepackage{aaai2027}  
\usepackage[hyphens]{url}  
\usepackage{graphicx} 
\usepackage{natbib}  
\usepackage{caption} 
\usepackage{algorithm}
\usepackage{algorithmic}

\usepackage{newfloat}
\usepackage{listings}
\DeclareCaptionStyle{ruled}{labelfont=normalfont,labelsep=colon,strut=off} 
\floatstyle{ruled}
\newfloat{listing}{tb}{lst}{}
\floatname{listing}{Listing}

\usepackage{booktabs}

\usepackage{multirow}
\usepackage{booktabs}
\usepackage{xcolor}
\usepackage{colortbl}
\usepackage{subcaption}
\usepackage{amsmath}
\usepackage{amssymb}
\usepackage{tabularx}
\usepackage{xspace}
\usepackage[capitalize]{cleveref}
\crefname{section}{Sec.}{Secs.}
\Crefname{section}{Section}{Sections}
\Crefname{table}{Table}{Tables}
\crefname{table}{Tab.}{Tabs.}
\Crefname{algorithm}{Algorithm}{Algorithms}

\makeatletter
\DeclareRobustCommand\onedot{\futurelet\@let@token\@onedot}
\def\@onedot{\ifx\@let@token.\else.\null\fi\xspace}
\makeatother

\newcommand{\method}{\textsc{MVHOI}\xspace}

\definecolor{best}{rgb}{0.96, 0.57, 0.58}
\definecolor{second}{rgb}{0.98, 0.78, 0.57}
\definecolor{third}{rgb}{1.0, 1.0, 0.56}

\makeatletter
\gdef\@reinserts{%
  \ifvoid\footins\else\insert\footins{\unvbox\footins}\fi
  \ifvoid\aaai@thanksins\else\insert\aaai@thanksins{\unvbox\aaai@thanksins}\fi
  \ifvoid\aaai@copyrightins\else\insert\aaai@copyrightins{\unvbox\aaai@copyrightins}\fi
  \ifvbox\@kludgeins\insert\@kludgeins{\unvbox\@kludgeins}\fi
}
\makeatother

\title{MVHOI: Bridge Multi-View Condition to Complex Human-Object Interaction Video Reenactment via 3D Foundation Model}

\author{
  Jinguang Tong\textsuperscript{\rm 1,\rm 2,\rm 3}\equalcontrib,
  Jinbo Wu\textsuperscript{\rm 3}\equalcontrib,
  Kaisiyuan Wang\textsuperscript{\rm 3},
  Zhelun Shen\textsuperscript{\rm 3},
  Xuan Huang\textsuperscript{\rm 4},\\
  Mochu Xiang\textsuperscript{\rm 3},
  Xuesong Li\textsuperscript{\rm 1},
  Yingying Li\textsuperscript{\rm 3},
  Haocheng Feng\textsuperscript{\rm 3},
  Chen Zhao\textsuperscript{\rm 3},\\
  Hang Zhou\textsuperscript{\rm 3},
  Wei He\textsuperscript{\rm 3},
  Chuong Nguyen\textsuperscript{\rm 2},
  Jingdong Wang\textsuperscript{\rm 3},
  Hongdong Li\textsuperscript{\rm 1},
}
\affiliations{
  \textsuperscript{\rm 1}The Australian National University\qquad
  \textsuperscript{\rm 2}CSIRO \qquad
  \textsuperscript{\rm 3}Baidu Inc. \qquad
  \textsuperscript{\rm 4}Sun Yat-sen University
}

\begin{document}

\maketitle

\begin{abstract}
  Human–Object Interaction (HOI) video reenactment aims to transfer the interaction dynamics of a source video to a novel target object while preserving realistic hand–object coordination. Existing methods typically rely on sparse 2D motion controls and monocular references, which are insufficient for complex out-of-plane motion and large viewpoint changes.
  We present \textbf{MVHOI}, a two-stage framework combining implicit motion extraction, 3D-aware multi-view reasoning, and video generation. In the first stage, a motion extractor encodes object dynamics into implicit motion descriptors. Conditioned on these descriptors, our Motion-Driven Object Prior (MDOP) module queries a 3D foundation model over multi-view references of the target object and autoregressively predicts coarse object anchors, a sequence of images that track the object's evolving orientation and appearance under the source motion without any explicit pose estimation. In the second stage, a DiT-based video generation model uses these anchors as structural guidance and the multi-view references as appearance guidance. We further reuse cross-view attention from MDOP as a soft attention bias to reduce reference-view confusion. For long videos, a cross-iterative inference strategy refreshes subsequent object priors using refined video outputs. Experiments demonstrate consistent improvements over state-of-the-art methods in object fidelity, motion consistency, visual quality, and interaction realism. Project page: \url{https://mvhoi.hirotong.fun}.
\end{abstract}

\begin{figure*}[t]
  \centering
  \includegraphics[width=\textwidth]{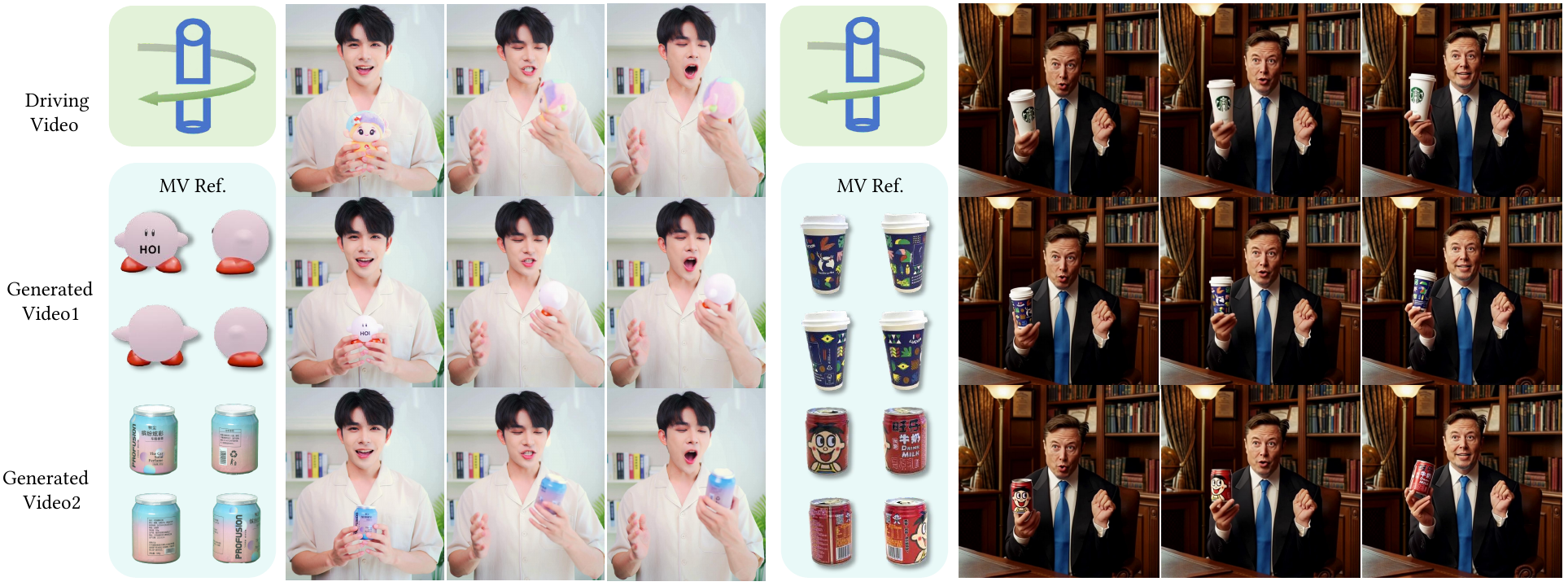}
  \caption{Given a source interaction video and multi-view images of a target object, MVHOI replaces the interacted object while faithfully following the source motion, preserving the target's identity and structure even under large rotations and viewpoint changes.}
  \label{fig:teaser}
\end{figure*}

\section{Introduction}
\label{sec:intro}

Reference-based human–object interaction (HOI) video reenactment aims to replace the interacted object in a source video with a novel target object while preserving the person, scene, and interaction dynamics (\cref{fig:teaser}), enabling applications such as product demonstration and virtual advertising. Despite the strong generative capacity of recent Video Foundation Models (VFMs)~\cite{wan2025,hunyuanvideo,cogvideox}, text or appearance conditions alone do not specify how the target object should evolve throughout the interaction, and reference-conditioned variants~\cite{hu2025hunyuancustom,jiang2025vace,phantom,kaleido} still fail to preserve the object's size and appearance under precise manipulation.

Many reference-based HOI methods~\cite{fan2025rehold,homa,hoi-swap,wang2025dreamactorh1} describe object motion with sparse 2D control signals (e.g., bounding boxes or keypoints) and represent the target object with a monocular reference image. While effective for image-plane motion, these conditions become ambiguous under non-planar motion and large viewpoint changes: \textbf{(1) Ambiguous motion representation:} sparse 2D signals cannot determine the object's evolving view-dependent state, leaving the video model to infer complex 3D motion from insufficient evidence; and \textbf{(2) Ambiguous multi-view appearance conditioning:} a monocular reference provides incomplete appearance for unseen viewpoints, while naively supplying multiple references does not establish which view matches the object state in each frame. Consequently, the generated object may exhibit structural distortion or appearance drift as its orientation changes.

To address these challenges, we introduce MVHOI, a two-stage framework built on the spatial priors of 3D Foundation Models (3DFMs). Our core insight is that a 3DFM can consolidate sparse multi-view inputs into a unified latent object representation that jointly encodes geometry and appearance across viewpoints, allowing HOI reenactment to be reformulated as a motion-conditioned querying process over this representation.

In the first stage, our \textbf{Motion-Driven Object Prior (MDOP)} module encodes the temporal dynamics of the source object into implicit motion latents, which then query a 3DFM that consolidates the multi-view references into a unified target-object representation. This produces a coarse yet view-consistent target-object reenactment sequence that provides motion-aligned structural guidance for subsequent video generation, without explicit 6D pose estimation or offline tracking.

In the second stage, a video generation model synthesizes the high-fidelity HOI video conditioned on the coarse sequence and the multi-view references. Specifically, we reuse the internal attention responses produced by MDOP as a soft correspondence between the moving object and the reference views, and design an inference-time attention enhancement mechanism that modulates their contributions accordingly. The coarse reenactment sequence thus provides motion-aware structural guidance, while the reused attention cues improve appearance consistency across viewpoint changes, coupling the two stages through the same 3D-aware prior.

Our contributions are summarized as follows: \textbf{1)} We introduce MVHOI, a two-stage framework that connects implicit source-object motion extraction, 3D-aware multi-view fusion, and video generation for cross-object HOI reenactment under complex non-planar motion and large viewpoint changes. \textbf{2)} We propose the MDOP module, which encodes source-object transformations as implicit motion latents and transfers them to the target object via motion-queried multi-view fusion within a 3D foundation model, producing coarse object anchors that provide target-specific structural guidance without explicit 6D pose estimation. \textbf{3)} We reuse the cross-view attention responses of MDOP to coordinate multi-view appearance conditioning in the video generator and introduce a cross-iterative inference strategy to reduce structural and appearance drift in long-video generation.

\section{Related Work}
\label{sec:related}
\subsection{Controllable Video Generation and Motion Transfer}
Early video generation methods~\cite{tune_a_video,svd,animatediff} adapted image diffusion models~\cite{ldm} to video, and transformer-based models~\cite{sora,cogvideox,hunyuanvideo,wan2025} scale up this paradigm with greatly improved temporal coherence and fidelity. Controllable variants preserve subject and object appearance through reference images or auxiliary conditioning branches~\cite{hu2025hunyuancustom,jiang2025vace,phantom}, while motion-transfer methods such as DisMo~\cite{ressler2025dismo} learn abstract motion representations that transfer across appearances and categories. However, these general-purpose methods do not associate source interaction dynamics with the multi-view geometry of a novel target object, and thus struggle to preserve object structure under substantial out-of-plane reorientation.

\subsection{HOI Video Generation and Reenactment}
Conditional HOI generation synthesizes interactions from predefined controls such as human poses, object trajectories, bounding boxes, meshes, or hand-object motion layouts~\cite{xu2024anchorcrafter,homa,wang2025dreamactorh1,pang2025manivideo}, and geometry-aware approaches further inject spatial priors as explicit 3D representations~\cite{chen2026hvg} or hierarchical multi-view geometry and texture features~\cite{xu2026geohoi}. These methods generate from predefined control signals rather than transferring interaction dynamics from a source video to a novel target object.

HOI reenactment methods instead replace the manipulated object within an existing interaction, via single-frame replacement with sequential warping~\cite{hoi-swap}, adaptive layout conditions~\cite{fan2025rehold}, two-stage inpainting~\cite{shen2025idit}, or temporally balanced and spatially selective reference injection~\cite{huang2026genhoi}. However, none of these methods constructs a time-varying target-object prior that couples the source interaction motion with the target object's multi-view geometry.


Recent 3D foundation models~\cite{dust3r,vggt,lin2025da3} learn transferable multi-view geometric representations, but are mostly used as static reconstruction or depth priors. In contrast, the 3D prior in MVHOI is motion-conditioned and dynamically aligned with the source interaction: source-motion latents and the target multi-view representation jointly produce a temporally varying object prior, whose geometry-aware attention cues additionally guide the video generation model.

\section{Method}
\label{sec:method}

\begin{figure*}[t]
  \centering
  \includegraphics[width=\linewidth]{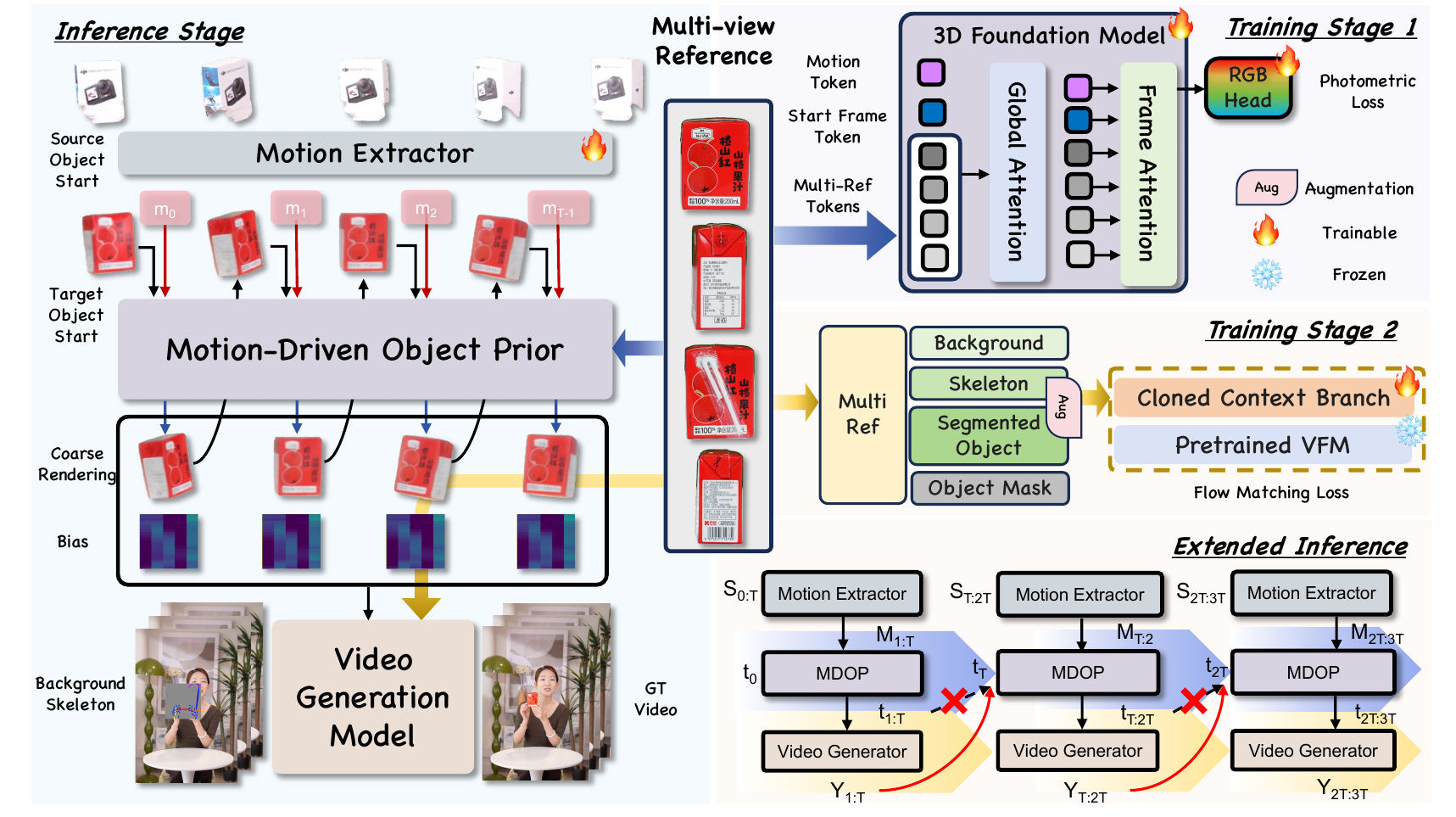}
  \caption{
    Overview of MVHOI. Given a source interaction video and multi-view images of a target object, the Motion-Driven Object Prior (MDOP) module converts implicit source-object dynamics into a sparse sequence of target-specific object anchors within a pretrained 3D-aware multi-view feature space. These anchors provide motion-aligned structural guidance to a DiT-based video generator, while the original multi-view references supply complementary appearance details. For long videos, the refined object state from one clip initializes the prior construction of the next clip.
  }
  \label{fig:pipeline}
\end{figure*}

\subsection{Overview}

We address cross-object Human--Object Interaction (HOI) video reenactment. Given a source interaction video $\mathcal{V}_{src}=\{X_t\}_{t=1}^{T}$ and $N$ reference images $\mathcal{I}_{ref}=\{I_i\}_{i=1}^{N}$ depicting a target object from different viewpoints, our goal is to synthesize $\mathcal{V}_{syn}=\{Y_t\}_{t=1}^{T}$, which should preserve the interaction dynamics of the source video while replacing the originally interacted object with the target object and maintaining the target's structure and appearance under substantial viewpoint changes.

As illustrated in \cref{fig:pipeline}, MVHOI follows a two-stage pipeline. In the first stage (Stage I), a motion extractor encodes the temporal dynamics of the source object into a sequence of implicit motion latents. These latents are fed into our \textbf{Motion-Driven Object Prior (MDOP)} module, which builds on a pretrained 3D foundation model to convert them into motion queries and jointly process the current target-object state and its multi-view references. MDOP then autoregressively predicts a sparse sequence of coarse target-object anchors that capture the evolving object state without explicit 6D pose estimation or 3D reconstruction. In the second stage (Stage II), a multi-reference video generation model uses the coarse anchors as structural guidance and the multi-view references as appearance guidance to synthesize the final HOI video with fine-grained object details. During long-video inference, the two components are further coupled across clips to reduce accumulated structural and appearance drift.

\begin{figure*}[t]
  \centering
  \includegraphics[width=\linewidth]{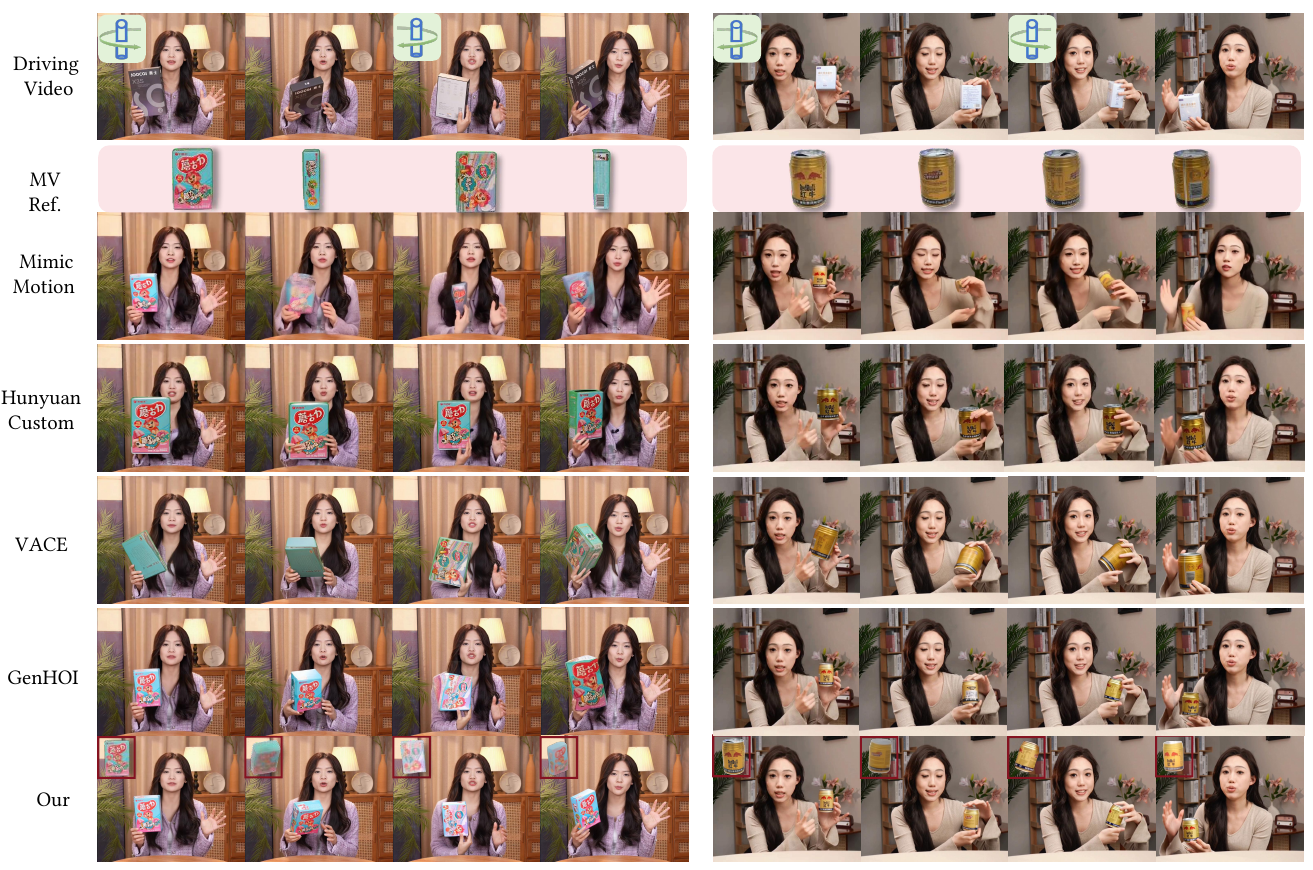}
  \caption{
    Qualitative comparison with state-of-the-art methods. Compared to MimicMotion~\cite{zhang2025mimicmotion}, HunyuanCustom~\cite{hu2025hunyuancustom}, VACE~\cite{jiang2025vace}, and GenHOI~\cite{huang2026genhoi}, our method generates object motion that more faithfully follows the driving video. Moreover, MVHOI better preserves the identity and structural consistency of the target object during complex manipulation. Detailed experimental settings are provided in \Cref{sec:experiments}.
  }
  \label{fig:main_compare}
\end{figure*}

\subsection{Motion-Driven Object Prior Construction}
The objective of MDOP is to transfer the source-object motion to the target object and predict the evolving visual states of the target object. Instead of relying on the sparse explicit motion signals used in prior works~\cite{xu2024anchorcrafter, wang2025dreamactorh1}, which become ambiguous under out-of-plane rotations, we use implicit motion descriptors extracted from the source-object sequence to condition the target-object state prediction.

\paragraph{Implicit source-motion encoding.}

We first extract the source-object crops $\mathcal{S}=\{S_t\}_{t=1}^{T}$ using corresponding object masks. We adapt the motion encoder $\mathcal{M}_{\theta}$ from DisMo~\cite{ressler2025dismo} to obtain a sequence of implicit motion latents $\mathbf{M}=\mathcal{M}_{\theta}(\mathcal{S})=\{m_t\}$ with $m_t\in\mathbb{R}^{d_m}$, computed at a temporal stride of $\Delta=4$ frames. Unlike explicit motion representations such as 6D poses or keypoints, \(m_t\) directly captures how the source object changes within a short video clip. This avoids the need for an additional pose estimation or pose fitting step.

\paragraph{Motion injection into the 3D foundation model.}
Although the implicit motion latent $m_t$ captures the source-object dynamics from time $t$ to $t+\Delta$, it does not directly specify how the target object should evolve under this transformation.
We therefore introduce, inside the 3D foundation model, a learnable motion-condition feature adapter $\mathcal{A}_{\phi}$ that projects the source-motion latent into a motion query based on the current target-object state $\hat{O}_t \in \mathbb{R}^{3\times H\times W}$. An image encoder $\mathcal{E}$ extracts a spatial feature map from $\hat{O}_t$, a modulation layer $\mathcal{H}$ maps $m_t$ to channel-wise scale and shift parameters $\gamma_t,\beta_t\in\mathbb{R}^{C}$, and a decoder $\mathcal{D}$ maps the modulated feature to the motion query:
\begin{equation}
  P_t
  =
  \mathcal{D}\!\left(
    \gamma_t \odot \operatorname{RMSNorm}\!\big(\mathcal{E}(\hat{O}_t)\big) + \beta_t
  \right),
  \label{eq:motion_query}
\end{equation}
with $[\gamma_t;\beta_t]=\mathcal{H}(m_t)$, where $C$ is the feature-channel dimension, $[\cdot;\cdot]$ denotes channel-wise concatenation, $\odot$ denotes element-wise multiplication, and $\gamma_t$ and $\beta_t$ are broadcast over the spatial dimensions. The three components together constitute the learnable adapter $\mathcal{A}_{\phi}$; its architectural details are provided in the Appendix.

Importantly, $P_{t}
\in\mathbb{R}^{3\times H\times W}$, which encodes the transition from $t$
to $t+\Delta$, is neither an explicit pose map nor the
predicted target-object frame at $t+\Delta$; it is patchified as an
additional query view for the subsequent 3D-aware reasoning process
described below.

\paragraph{3D-aware multi-view fusion and anchor decoding.}

Recent feed-forward 3D foundation models learn transferable cross-view representations from multiple visual observations~\cite{vggt,lin2025da3}. We instantiate our multi-view fusion backbone
$\mathcal{G}_{\psi}$ from Depth Anything 3~\cite{lin2025da3}. At each sampled time step, it jointly processes the current target-object state, the $N$ reference images, and the motion query:
\begin{equation}
  Z_t
  =
  \mathcal{G}_{\psi}
  \left(
    [\hat{O}_t,I_1,\ldots,I_N,P_t]
  \right).
\end{equation}
The backbone alternates intra-view and cross-view attention, enabling the motion query to aggregate the structure and appearance of the target object from its reference views.
The resulting joint tokens serve as the internal representation for anchor prediction rather than an explicit reconstruction of the target object in 3D space.

We extract the features associated with the motion-query view, $Z_{t,P}$, and decode the next coarse object state with an RGB prediction head, $\hat{O}_{t+\Delta}=\mathcal{R}(Z_{t,P})$; in practice $\mathcal{R}$ consumes features from several intermediate layers of $\mathcal{G}_{\psi}$ (see Appendix).
Starting from an initial target-object state $\hat{O}_0$, MDOP applies this process autoregressively to obtain a sparse sequence of coarse object anchors $\hat{\mathcal{O}}=\{\hat{O}_0,\hat{O}_{\Delta},\hat{O}_{2\Delta},\ldots\}$. Although these anchors do not preserve all high-frequency texture details, they encode the target object's evolving orientation, silhouette, and coarse appearance under the source motion. 

\subsection{Video Generation with Multi-Reference Conditioning}

Given the sparse object anchors $\hat{\mathcal{O}}$ produced by MDOP, we employ a DiT-based video inpainting model~\cite{wan2025} to synthesize the final HOI video. The coarse anchors and multi-view references provide complementary conditions: the anchors constrain the frame-dependent object state, whereas the reference images provide high-quality appearance information that is absent from the coarse predictions.

\paragraph{Coarse-prior and reference conditioning.}

To inject the two conditions into the video-generation backbone, we adopt an additional context branch following VACE~\cite{jiang2025vace}. The target-object reference images are prepended to the context sequence and processed by the cloned context branch, preserving their photometric information. Meanwhile, the temporally aligned coarse anchors are incorporated into the masked video stream using the HOI masks. Self-attention then propagates information among the noisy video tokens, coarse structural conditions, and multi-view reference tokens.

Directly optimizing this component with MDOP predictions would require repeated inference of the 3D foundation model throughout training, leading to prohibitive computational costs. We therefore construct proxy guidance from the ground-truth object crops. Naively using clean object crops, however, would cause appearance leakage and allow the video model to ignore the multi-view references. To approximate the degraded characteristics of MDOP predictions, we apply geometric perturbations, photometric jittering, blur, and noise to the proxy guidance. These augmentations suppress high-frequency appearance information while preserving coarse shape and orientation, encouraging the video model to use the object anchors for structural alignment and the reference images for detailed appearance synthesis.

\paragraph{Inference-time attention enhancement.}

Although multi-view reference tokens provide complementary appearance information, appearance-driven attention may assign excessive weight to reference views that are incompatible with the current object state. Inspired by prior studies showing that diffusion generation can be controlled through attention manipulation~\cite{hertz2022prompt}, we use the cross-view associations formed inside MDOP to modulate reference conditioning during inference.

Let $\bar{A}_{t,i}$ denote the mean attention between the motion-query tokens and the tokens of the $i$-th reference view, averaged over all heads and token pairs at a designated layer of the MDOP fusion backbone $\mathcal{G}_{\psi}$. We normalize these associations into a frame-to-view weight and convert it into a reference-attention bias:
\begin{equation}
  w_{t,i}
  =
  \frac{\bar{A}_{t,i}}{\sum_{i'=1}^{N}\bar{A}_{t,i'}},
  \qquad
  B_{t,i}
  =
  \alpha
  \log
  \frac{w_{t,i}+\epsilon}
  {1-w_{t,i}+\epsilon},
\end{equation}
where $\alpha$ controls the bias strength and $\epsilon$ ensures numerical stability. Rather than treating $w_{t,i}$ as a hard retrieval decision, we use it to softly modulate the relative influence of different reference views: the bias is broadcast to the reference tokens associated with view $i$ and added to the corresponding attention logits:
\begin{equation}
  \label{eq:attention_bias}
  \operatorname{Attn}(Q,K,V;B)
  =
  \operatorname{softmax}
  \left(
    \frac{QK^{\top}}{\sqrt{d}}+B
  \right)V.
\end{equation}

\begin{table*}[ht]
  \centering
  \small
  \setlength{\tabcolsep}{4pt} 
  \begin{tabular}{l|cccccc|ccc}
    \toprule
    \multirow{2}{*}{Method} & \multicolumn{6}{c|}{Self-Reenactment} & \multicolumn{3}{c}{Cross-Reenactment} \\
    \cmidrule{2-10}
    & PSNR $\uparrow$ & SSIM $\uparrow$ & LPIPS $\downarrow$ & FID $\downarrow$ & FVD $\downarrow$ & O-CLIP $\uparrow$ & FID $\downarrow$ & FVD $\downarrow$ & O-CLIP $\uparrow$ \\
    \midrule
    MimicMotion & 19.34 & 0.806 & 0.233 & 56.13 & 632.9 & 0.685 & 84.52 & 774.2 & 0.470 \\
    VACE & 27.25 & \underline{0.954} & 0.036 & 47.43 & 204.6 & 0.706 & 62.54 & 308.1 & 0.556 \\
    HuMo & 8.71 & 0.497 & 0.716 & 231.03 & 2120.8 & 0.449 & 225.81 & 2282.5 & 0.322 \\
    HunyuanCustom & 25.56 & 0.925 & 0.090 & 48.88 & 247.4 & 0.791 & 62.68 & 349.7 & 0.640 \\
    GenHOI & \underline{29.60} & 0.949 & \underline{0.035} & \underline{19.24} & \textbf{76.8} & \textbf{0.866} & \underline{48.01} & \underline{280.4} & \textbf{0.680} \\
    \midrule
    Ours & \textbf{30.21} & \textbf{0.960} & \textbf{0.025} & \textbf{17.90} & \underline{86.6} & \underline{0.848} & \textbf{41.14} & \textbf{274.0} & \underline{0.645} \\
    \bottomrule
  \end{tabular}
  \caption{Quantitative comparison with baseline methods on Self-Reenactment and Cross-Reenactment. Our method achieves the best results on most metrics, with clear advantages in reconstruction fidelity, perceptual quality, and temporal consistency. Best and second-best results are highlighted in \textbf{bold} and \underline{underline}, respectively.}
  \label{tab:exp_quantitative}
\end{table*}

This operation increases the contribution of reference tokens whose views are more compatible with the current coarse object state. We therefore refer to it as \emph{inference-time attention enhancement} (AE), rather than explicit appearance retrieval. The extracted associations shift smoothly across reference views as the object orientation evolves (visualized in the Appendix), indicating that MDOP encodes a meaningful view-selection signal.

\subsection{Training Objectives}

As discussed above, we train MDOP and the multi-reference video generator separately for computational efficiency, and integrate them only at inference time.

\paragraph{Motion-driven object prior construction.}

We train MDOP to predict coarse target-object anchors that preserve the source object's motion dynamics while maintaining the target object's appearance. These predicted anchors are supervised by
\begin{equation}
\mathcal{L}_{\mathrm{MDOP}}=\lambda_{1}\mathcal{L}_{2}+\lambda_{2}\mathcal{L}_{\mathrm{LPIPS}}+\lambda_{3}\mathcal{L}_{\mathrm{SSIM}},
\end{equation}
combining pixel reconstruction, perceptual similarity, and structural similarity terms.

\paragraph{Multi-reference video generation.}

We train the video generator with the standard flow-matching objective~\cite{lipman2022flow_matching,liu2022rectified_flow},
\begin{equation}
\mathcal{L}_{\mathrm{FM}}=\mathbb{E}\big[\|v_{\vartheta}(x_t,t,c)-u_t\|_2^2\big],
\end{equation}
where $v_{\vartheta}$ is the predicted velocity field, $u_t$ is the target velocity, and $c$ denotes the conditioning inputs.
Following the HOI-region emphasis used in AnchorCrafter~\cite{xu2024anchorcrafter}, we assign a larger weight to the relatively small interaction region. Let
$\mathbf{M}_{\mathrm{HOI}}\in\{0,1\}$ denote its binary mask. We define the spatial weighting function as
\begin{equation}
    \rho(j)
    =
    1+
    \left(
    \eta\frac{S}{S_{\mathrm{HOI}}}-1
    \right)
    \mathbf{M}_{\mathrm{HOI}}(j),
\end{equation}
where $S$ and $S_{\mathrm{HOI}}$ are the areas of the full image and the HOI region, respectively. The final video objective is
\begin{equation}
    \mathcal{L}_{\mathrm{video}}
    =
    \mathbb{E}
    \left[
    \frac{1}{S}
    \sum_{j}
    \rho(j)
    \left\|
    v_{\vartheta}(x_t,t,c)_j-u_t(j)
    \right\|_2^2
    \right],
\end{equation}
where the factor $S/S_{\mathrm{HOI}}$ compensates for the small spatial extent of the interaction region, and $\eta$ controls its relative importance.

\subsection{Cross-Iterative Long-Video Inference}

Long-video generation requires maintaining target-object identity and motion stability over multiple temporal segments. Independently rolling out coarse anchors over the entire sequence can accumulate structural errors, whereas continuously extending the video generator from its previous predictions may cause appearance drift.

We therefore couple the two components across successive clips. For temporal segment $s$, MDOP first predicts a set of coarse object anchors $\hat{\mathcal{O}}^{(s)}$. The video generator then synthesizes a refined clip $\mathcal{Y}^{(s)}$ using the anchors and target-object references. Instead of initializing the next MDOP rollout from its previous coarse prediction, we extract the target-object state from the final refined frame of $\mathcal{Y}^{(s)}$, $\hat{O}^{(s+1)}_0=\operatorname{Crop}_{obj}(Y^{(s)}_{\mathrm{end}})$.
This cross-iterative loop lets MDOP repeatedly provide motion-aligned structural constraints to the video generator, while the refined video output refreshes the visual state used by MDOP, mitigating the accumulation of coarse-prediction errors and long-term appearance drift.

\section{Experiments}
\label{sec:experiments}

\subsection{Implementation Details}
We initialize the implicit motion extractor from the pretrained DisMo~\cite{ressler2025dismo}. For the MDOP module, we construct it from a pretrained 3D foundation model~\cite{lin2025da3} with the proposed motion-condition adapter and train on a synthetic cross-object dataset rendered from Objaverse~\cite{deitke2023objaverse} and a self-collected dataset ($\sim$100 hours) with multi-view reference images. As for the video generation model, we initialize it from the pretrained Wan2.1-T2V-14B~\cite{wan2025} with the additional context branch and train it on the same self-collected dataset. More details about data collection, preprocessing, and training are provided in the Appendix.

\subsection{Evaluation Setup}
We evaluate MVHOI under two settings: \textbf{Self-Reenactment}, which reconstructs HOI videos from the masked source video and references depicting the same object, and \textbf{Cross-Reenactment}, which replaces the interacted object with a different one. All source and target pairs are fixed in advance and shared by all methods.

\paragraph{Metrics.}
We report FID~\cite{heusel2017fid} and FVD~\cite{unterthiner2018fvd} to assess video quality, and Object-CLIP (O-CLIP)~\cite{xu2024anchorcrafter,homa} for target-object fidelity. For Self-Reenactment, where frame-aligned ground truth is available, we additionally report PSNR~\cite{al2024psnr}, SSIM~\cite{wang2004ssim}, and LPIPS~\cite{zhang2018lpips} for reconstruction fidelity. Cross-Reenactment is further evaluated with a human study of motion consistency (MC), visual quality (VQ), and HOI realism (HR); exact metric implementations and the study protocol are described in the Appendix.

\paragraph{Baselines.}
We compare \method with MimicMotion~\cite{zhang2025mimicmotion}, VACE~\cite{jiang2025vace}, HunyuanCustom~\cite{hu2025hunyuancustom}, HuMo~\cite{chen2025humo}, and GenHOI~\cite{huang2026genhoi}, spanning pose-guided human animation, unified controllable video editing, multi-subject reference-conditioned generation, and HOI reenactment. All methods are evaluated with their official checkpoints, except VACE, which we fine-tune on our collected dataset with multiple reference views for a fair comparison. Per-method input adaptations are detailed in the Appendix.

\subsection{Experimental Results}
\label{sec:experimental_results}

\begin{figure}[t]
  \centering
  \includegraphics[width=\linewidth]{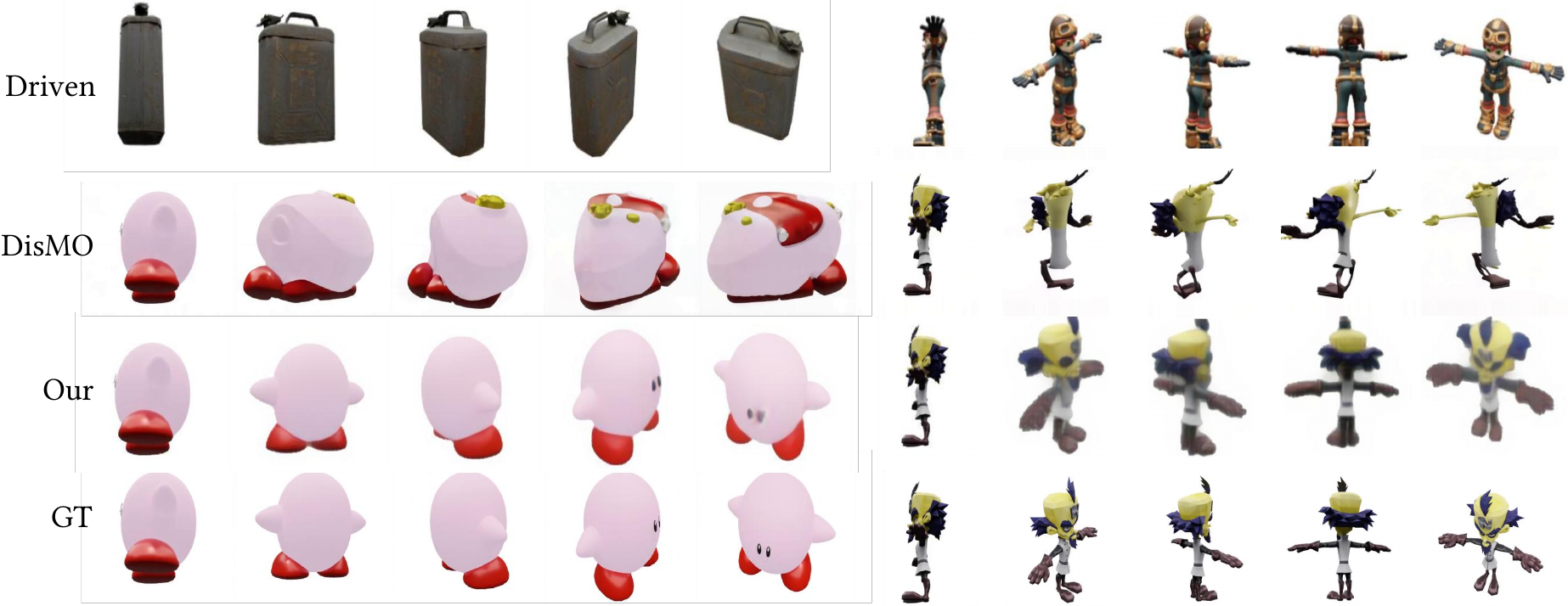}
  \caption{
    Qualitative comparison of cross-object motion transfer. The source and target are different objects. While DisMo~\cite{ressler2025dismo} transfers motion from a single target-object initialization, MDOP additionally reasons over multi-view target-object observations, leading to improved structural and appearance consistency under viewpoint changes.
  }
  \label{fig:dismo_compare}
\end{figure}

\paragraph{Evaluation of motion-driven object priors.}
We first evaluate MDOP in isolation from the video generator on a cross-object transfer task, where both MDOP and DisMo~\cite{ressler2025dismo} receive the same pretrained motion representation, source motion, and target-object initialization; any gain is therefore attributable to target-aware multi-view fusion. As shown in \cref{tab:dismo}, MDOP outperforms DisMo in both reconstruction fidelity and perceptual quality, and the qualitative results in \cref{fig:dismo_compare} show correspondingly better structural and appearance consistency under viewpoint changes. These results support our central design: coupling implicit source motion with multi-view observations of the target object yields a more target-specific and structurally stable dynamic prior.

\paragraph{Full HOI video reenactment.}
We evaluate the complete MVHOI framework under Self-Reenactment and Cross-Reenactment, and report quantitative results in \cref{tab:exp_quantitative}. MVHOI achieves the best scores on most metrics across both settings: under Self-Reenactment it attains the best reconstruction fidelity and perceptual quality, accurately recovering real HOI videos from multi-view references of the same product, while under the more challenging Cross-Reenactment setting it obtains the best FID and FVD, indicating improved preservation of the source dynamics and stronger overall video quality. GenHOI~\cite{huang2026genhoi} obtains higher O-CLIP scores, reflecting its strength in direct reference-appearance preservation, whereas MVHOI provides better distribution-level video quality and motion preservation by explicitly constraining the evolving object state through its motion-driven 3D-aware prior. The qualitative results in \cref{fig:main_compare} support these numbers. MVHOI faithfully follows the source motion while preserving the target object's identity fidelity and multi-view appearance consistency under large movements, where the baselines often exhibit unstable object shape or drifting appearance.

\begin{table}[t]
  \centering
  \small
  \begin{tabular}{l|cccc}
    \toprule
    Method
    & PSNR $\uparrow$
    & SSIM $\uparrow$
    & FID $\downarrow$
    & KID ($10^{-3}$) $\downarrow$ \\
    \midrule
    DisMo
    & 16.56
    & 0.8081
    & 111.96
    & 24.81 \\
    Ours
    & \textbf{20.75}
    & \textbf{0.8304}
    & \textbf{84.58}
    & \textbf{9.67} \\
    \bottomrule
  \end{tabular}
  \caption{
    Quantitative comparison with DisMo on held-out cross-object pairs from Objaverse. MDOP improves both frame-aligned reconstruction fidelity and perceptual distribution quality.
  }
  \label{tab:dismo}
\end{table}

\begin{table}[t]
  \centering
  \small
  \setlength{\tabcolsep}{4pt}
  \begin{tabular}{l|ccc|ccc}
    \toprule
    \multirow{2}{*}{Method} & \multicolumn{3}{c|}{Short} & \multicolumn{3}{c}{Long} \\
    \cmidrule{2-7}
    & MC $\uparrow$  & VQ  $\uparrow$ & HR $\uparrow$ & MC $\uparrow$  & VQ  $\uparrow$ & HR $\uparrow$  \\
    \midrule
    VACE             & 3.66 & 3.20 & 3.32 & 3.73 & 2.14 & 3.27\\
    HunyuanCustom   & 3.59 & 3.29 & 3.11 & 3.49 & 3.56 & 3.17\\
    GenHOI          & 4.20 & 4.10 & 4.15 & 4.07 & 3.94 & 3.89\\
    Ours &  \textbf{4.47} & \textbf{4.34} & \textbf{4.34} & \textbf{4.50} & \textbf{4.32} & \textbf{4.32} \\
    \bottomrule
  \end{tabular}
  \caption{User study on Cross-Reenactment for short (single-clip) and long (10-second) video generation. MC, VQ, and HR denote motion consistency, visual quality, and HOI realism, rated on a 1--5 scale (higher is better).}
  \label{tab:user_study}
\end{table}

\paragraph{User study.}
\Cref{tab:user_study} reports the human evaluation on Cross-Reenactment for both short and long video generation. Our method receives clearly higher visual-quality ratings than all baselines, together with the best motion consistency and HOI realism. Moreover, MVHOI maintains its performance when moving from short to long video generation, whereas the ratings of competing methods degrade generally, demonstrating the effectiveness of our cross-iterative inference strategy for long-horizon stability. More qualitative video results are provided in the supplementary material.


\subsection{Ablation Study}
\label{sec:ablation_study}

\begin{table}[t]
  \centering
  \small
  \setlength{\tabcolsep}{4pt}
  \begin{tabular}{lccc}
    \toprule
    Method & FID $\downarrow$ & FVD $\downarrow$ & O-CLIP $\uparrow$ \\
    \midrule
    Baseline & 46.55 & 296.7 & 0.611 \\
    +MDOP & 45.12 & 282.1 & 0.642 \\
    +MDOP+AE (full) & \textbf{41.14} & \textbf{274.0} & \textbf{0.645} \\
    \bottomrule
  \end{tabular}
  \caption{Ablation study on Cross-Reenactment. MDOP denotes conditioning the video generator on the coarse object anchors from our Motion-Driven Object Prior module, and AE denotes the inference-time attention enhancement.}
  \label{tab:exp_ablation}
\end{table}

We conduct an ablation study to evaluate the contribution of two key components in our framework: the \textbf{coarse object anchors produced by MDOP} and the \textbf{inference-time attention enhancement (AE)}. As the baseline, we train a multi-reference adapter conditioned solely on multi-view reference images. Quantitative results on Cross-Reenactment are reported in \cref{tab:exp_ablation}, with qualitative comparisons shown in the Appendix.

Overall, both components contribute consistently to improved generation quality and temporal stability. Conditioning on the MDOP anchors provides explicit motion-aware structural guidance, which stabilizes object dynamics and reduces temporal inconsistency. Building on this, AE further improves appearance fidelity by encouraging the model to retrieve viewpoint-consistent details from the multi-view references, thereby alleviating view confusion and enhancing cross-frame appearance consistency.

\section{Conclusion}
\label{sec:conclusion}

We present \method, a framework that bridges multi-view object conditions and HOI video reenactment through 3D foundation models. The proposed Motion-Driven Object Prior (MDOP) module converts implicit source motion and multi-view references into motion-aligned structural guidance, an inference-time attention enhancement improves viewpoint-consistent appearance, and a cross-iterative inference strategy suppresses drift in long-video generation. Experiments across object reenactment, HOI video generation, and long-horizon synthesis validate explicit 3D-aware guidance coupled with multi-view appearance modeling, with consistent quantitative and perceptual gains.

\bibliography{main}

\appendix
\clearpage
\section{Implementation Details}
\label{app:implementation}

\subsection{Data Collection and Preprocessing}
\label{app:data}
We collect approximately 100 hours of real-world HOI videos, organized as 75K five-second clips and covering around 600 distinct products. Each video is associated with multiple product images depicting the interacted object from different viewpoints. We first use SAM2~\cite{ravi2024sam} to segment the interacted object throughout the video and extract a sequence of object crops. RTMW~\cite{jiang2024rtmw} is used to obtain human-pose conditions, and a vision-language model~\cite{yu2025minicpm} generates the corresponding video descriptions.

For the synthetic cross-object data, we construct training pairs from 40{,}000 Objaverse objects~\cite{deitke2023objaverse}. In each pair, the source and target correspond to different object instances. For each object, we render 20 motion frames; the source sequence provides the driving motion, while the target object is rendered under the corresponding motion trajectory to provide supervision. We additionally render four reference images of the target object from distinct viewpoints. Consequently, the source motion and target appearance cannot be matched through identity copying: the model must transfer the motion encoded from the source sequence to the geometrically different target object by reasoning over its multi-view observations.

\subsection{Network Architecture Details}
\label{app:adapter}
\paragraph{Motion-condition feature adapter.}
The adapter is a lightweight convolutional modulation module. The image encoder $\mathcal{E}$ uses two $3\times3$ convolutional layers with ReLU activations to map the current three-channel object state to a 128-channel feature map. The DisMo motion extractor produces a 128-dimensional motion embedding, which the linear modulation layer $\mathcal{H}$ maps to channel-wise scale and shift parameters. These parameters modulate the RMS-normalized image feature as defined in the main paper. The decoder $\mathcal{D}$ mirrors the encoder with two $3\times3$ convolutional layers and maps the modulated feature back to the three-channel motion query $P_t$. The motion extractor, $\mathcal{E}$, $\mathcal{H}$, and $\mathcal{D}$ are jointly optimized with MDOP.

\paragraph{RGB prediction head.}
Following the dense-prediction design of VGGT~\cite{vggt} and DA3-Large~\cite{lin2025da3}, the RGB head uses the same four intermediate backbone layers as DA3-Large, indexed by $\{11,15,19,23\}$. We retain the tokens associated with the motion-query view and decode them through the standard DPT multi-scale projection and coarse-to-fine residual fusion pathway. The fused feature is bilinearly upsampled to the input resolution and mapped to four output channels with a lightweight convolutional decoder. After a sigmoid activation, the first three channels form the predicted RGB anchor $\hat{O}_{t+\Delta}$, while the fourth is an auxiliary confidence channel; only the RGB prediction is used in the autoregressive anchor rollout.

\subsection{Training Details}
\label{app:training}

\paragraph{Stage-I training.}
All object crops and target reference images are processed at a resolution of $266\times266$. The implicit motion encoder is initialized from DisMo~\cite{ressler2025dismo}. The multi-view fusion backbone of the 3D foundation model is initialized from DA3-Large~\cite{lin2025da3}, while the motion-condition feature adapter, the adapted multi-view backbone, and the RGB prediction head are jointly optimized. We train these modules using a mixture of real-HOI self-reconstruction samples and synthetic Objaverse cross-object samples. Training is conducted on 8 NVIDIA A800 (80GB) GPUs for 10{,}000 iterations with a batch size of 64 and a fixed learning rate of $6\times10^{-4}$. The reconstruction-loss weights are set to $\lambda_1=1.0$, $\lambda_2=0.1$, and $\lambda_3=0.1$, respectively.

\paragraph{Stage-II training.}
The video-generation backbone is initialized from Wan2.1-T2V-14B~\cite{wan2025}. Following VACE~\cite{jiang2025vace}, we initialize the additional context branch by cloning the corresponding pretrained backbone blocks and train it on the collected real HOI dataset using the proxy-guidance construction described in the main paper. Stage II is trained on 81-frame clips at 720p resolution for 20{,}000 iterations on 16 NVIDIA A800 (80GB) GPUs with a per-GPU batch size of 1 (effective batch size 16), a learning rate of $1\times10^{-5}$, and bf16 mixed precision.

\paragraph{Inference details.}
Videos are generated with 50 sampling steps and a classifier-free guidance scale of 5.0. The object mask is expanded before conditioning, leaving room for target objects whose shape differs from the source. The attention enhancement operates only at inference and introduces no additional training cost. Long videos are generated in segments of up to 81 frames; quantitative and qualitative results are provided in \cref{app:long_video}.

\section{Evaluation Protocol}
\label{app:evaluation}

\subsection{Evaluation Datasets}
\label{app:eval_datasets}
We evaluate MVHOI on three complementary test sets covering synthetic cross-object transfer, real-world self-reconstruction, and real-world cross-object reenactment.

First, we construct a held-out synthetic evaluation set from Objaverse~\cite{deitke2023objaverse} to evaluate MDOP independently. The source and target objects in each pair have different identities, and none of the selected objects overlap with the Stage-I training set. Because the target object can be rendered under the same motion trajectory as the source object, this test set provides frame-aligned ground truth for evaluating cross-object motion transfer.

Second, we select 100 held-out HOI videos from our collected real-world dataset. Each video is accompanied by multi-view images of the product appearing in the interaction. These videos and products are excluded from the data used to train both stages. We use this subset for Self-Reenactment by conditioning the model on multi-view images of the same product and evaluating its reconstruction of the original HOI video.

Third, we evaluate real-world Cross-Reenactment using source videos paired with multi-view images of different target products. This test set contains 142 fixed source--target pairs: 29 representative source sequences from the publicly released AnchorCrafter dataset~\cite{xu2024anchorcrafter} and 113 pairs constructed from our held-out HOI data. The source and target identities are always different in this setting. Since AnchorCrafter provides three product views, all methods use the same three reference images on this subset. For our collected data, four reference views are used when available. The source--target pairs are fixed in advance and shared by all evaluated methods.

\subsection{Evaluation Metrics}
\label{app:metrics}
Generated and ground-truth videos are matched by sample identifier and decoded in RGB. For frame-aligned metrics, we resize the ground-truth frames to the spatial resolution of the generated video, using area interpolation for downsampling and bicubic interpolation for upsampling. If the two videos have different lengths, the ground-truth sequence is truncated to the generated length or padded by repeating its final frame. Unless otherwise stated, the reported score is the mean over all evaluated frames and videos.

\paragraph{Frame-aligned reconstruction metrics.}
We compute PSNR~\cite{al2024psnr} on full RGB frames in the $[0,255]$ range with a peak value of 255. SSIM~\cite{wang2004ssim} is computed on RGB frames with a data range of 255 using the channel-aware implementation from \texttt{scikit-image}. LPIPS~\cite{zhang2018lpips} uses the learned AlexNet backbone; RGB inputs are normalized from $[0,255]$ to $[-1,1]$ before feature extraction. These three metrics are used only when frame-aligned ground truth is available.

\paragraph{Distributional video-quality metrics.}
FID~\cite{heusel2017fid} compares the pooled distributions of all generated and corresponding real-video frames. We use the 2048-dimensional Inception-V3 representation implemented by \texttt{torchmetrics}, process RGB frames as 8-bit tensors, and compute the Fr\'echet distance from the empirical means and covariance matrices of the two frame pools. KID~\cite{binkowski2018demystifying}, used for the Stage-I synthetic Cross-Object Transfer experiment, is computed from the same Inception representation with subsets of 50 samples, and we report the estimated mean kernel distance.

For FVD~\cite{unterthiner2018fvd}, each video is decoded in its entirety, bicubically resized to $224\times224$, and normalized to $[-1,1]$. We extract one video-level representation with an I3D network pretrained on Kinetics-400 and compute the Fr\'echet distance between the generated- and real-video feature distributions. Generated and real videos are paired using the same fixed evaluation list, and all compared methods are evaluated with the same temporal extent.

\paragraph{Object fidelity.}
O-CLIP~\cite{xu2024anchorcrafter,homa} is computed with the OpenCLIP ViT-B/16 image encoder pretrained on LAION-2B (\texttt{laion2b\_s34b\_b88k}). For every generated frame, we use the corresponding ground-truth object mask, binarized at 127, to remove background pixels and derive a tight object bounding box. The crop is zero-padded to a square and bilinearly resized to $224\times224$. Both the crop and each available target-object reference view are encoded into $\ell_2$-normalized CLIP features. We compute their cosine similarities, select the maximum across reference views for each frame, and average these frame-level maxima over each video and then over the evaluation set. This maximum-over-views aggregation allows the metric to match each generated object state to its most compatible observed target view.

\subsection{Human Study Protocol}
\label{app:human_study}
In addition to automatic metrics, we conduct a human study along three axes: \textbf{Motion Consistency (MC)}, \textbf{Visual Quality (VQ)}, and \textbf{Human--Object Interaction Realism (HR)}. MC measures whether the synthesized object follows the driving motion and remains synchronized with the hand throughout manipulation (e.g., avoiding drift, slipping, or temporal jitter). VQ assesses overall perceptual quality, including sharpness, visible artifacts, and temporal smoothness. HR evaluates the physical plausibility and naturalness of the interaction, with particular emphasis on contact correctness and realistic hand--object coupling under occlusions and large viewpoint changes. The study involves 30 participants and covers all 142 short-video Cross-Reenactment pairs and 113 long-video test cases. Videos are presented in randomized order with method identities concealed. All 30 participants rate every generated video from each method included in the user study along all three axes on a 1--5 scale (5 indicates the best quality), and we report the mean score over all participants and test cases.

\subsection{Baseline Adaptation Details}
\label{app:baselines}
We adapt inputs according to each baseline's native conditioning interface, as specified below. All methods generate videos at matched spatial resolution, temporal length, and frame rate. Unless noted otherwise, we use official checkpoints and recommended inference settings. No test sequence or target product is used to train or fine-tune any evaluated model.

\paragraph{MimicMotion.}
MimicMotion is a pose-guided human animation method and does not natively support object replacement. To prepare its single-image input, we use Google's Nano Banana image-editing model (November 2025 version) in an offline preprocessing step. Given the first frame of the source interaction video and a target-object reference image, it edits the first frame to contain the target product. We then provide the edited first frame and the human-pose sequence extracted from the source interaction video to MimicMotion. Nano Banana is used only to prepare this initial condition and is not invoked during MimicMotion video generation; all subsequent frames are generated through MimicMotion's original pose-control pathway.

\paragraph{VACE.}
VACE is the closest controllable-editing baseline to our Stage-II setup. We feed it the same masked source video and available structural conditions used by our video-generation stage (e.g., depth and related structure maps when supported). Because the official checkpoint is not specialized for multi-view HOI reenactment with product references, we additionally fine-tune VACE on our collected multi-view HOI dataset with multiple reference views, under the same data split used for \method. This fine-tuning is applied only to VACE among the compared methods, so that the comparison reflects a strong domain-adapted controllable baseline rather than an out-of-domain official checkpoint.

\paragraph{HunyuanCustom.}
HunyuanCustom is evaluated as a multi-subject reference-conditioned generator. We provide the masked source video together with a front-view target-product image as the object reference, following its recommended subject-conditioning protocol.

\paragraph{HuMo.}
HuMo receives a human reference image, target-product reference images, and the corresponding text description. This matches its native multi-subject conditioning interface and allows the model to condition on both the interacting person and the replacement object.

\paragraph{GenHOI.}
GenHOI is an HOI-oriented reenactment baseline. We follow its official inference protocol and provide the source interaction video together with target-product references, without modifying its training setup.

\section{Additional Experiment Results}
\label{app:more_results}

\subsection{Cross-View Association Visualization}
\label{app:attnmap}

\begin{figure}[t]
  \centering
  \includegraphics[width=\linewidth]{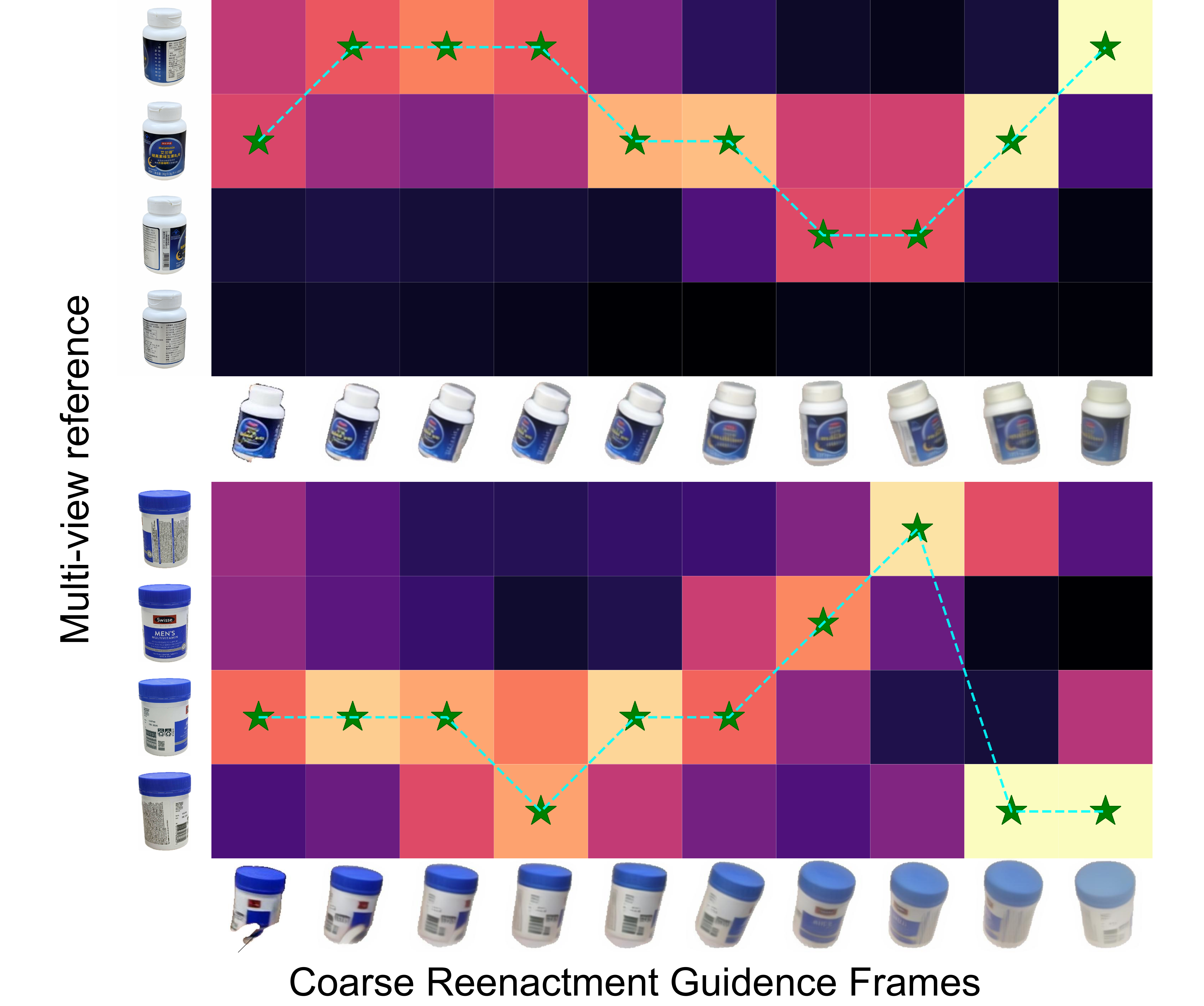}
  \caption{
    Cross-view association weights between motion-conditioned query states and target-object reference views.
  }
  \label{fig:exp_attnmap}
\end{figure}

Appearance-driven attention in a video diffusion model may favor a visually similar but geometrically incompatible reference, particularly during large rotations or hand--object occlusions. To examine the view-selection signal formed inside MDOP, \cref{fig:exp_attnmap} visualizes the normalized association weights between each motion-conditioned query state and the target-object reference views. Each row corresponds to a query state along the object trajectory, while the columns correspond to the available reference viewpoints.

As the visible side of the object changes, the attention mass shifts toward geometrically compatible reference views rather than remaining fixed on a single appearance reference. Intermediate object states also distribute weight across adjacent views instead of making an abrupt hard selection. This visualization suggests that MDOP captures a correspondence between the evolving object state and the reference viewpoints. We reuse these associations as a soft attention bias in Stage II, where they complement appearance similarity and reduce reference-view confusion.

\subsection{Qualitative Ablation Results}
\label{app:ablation_qualitative}

\begin{figure}[t]
  \centering
  \includegraphics[width=\linewidth]{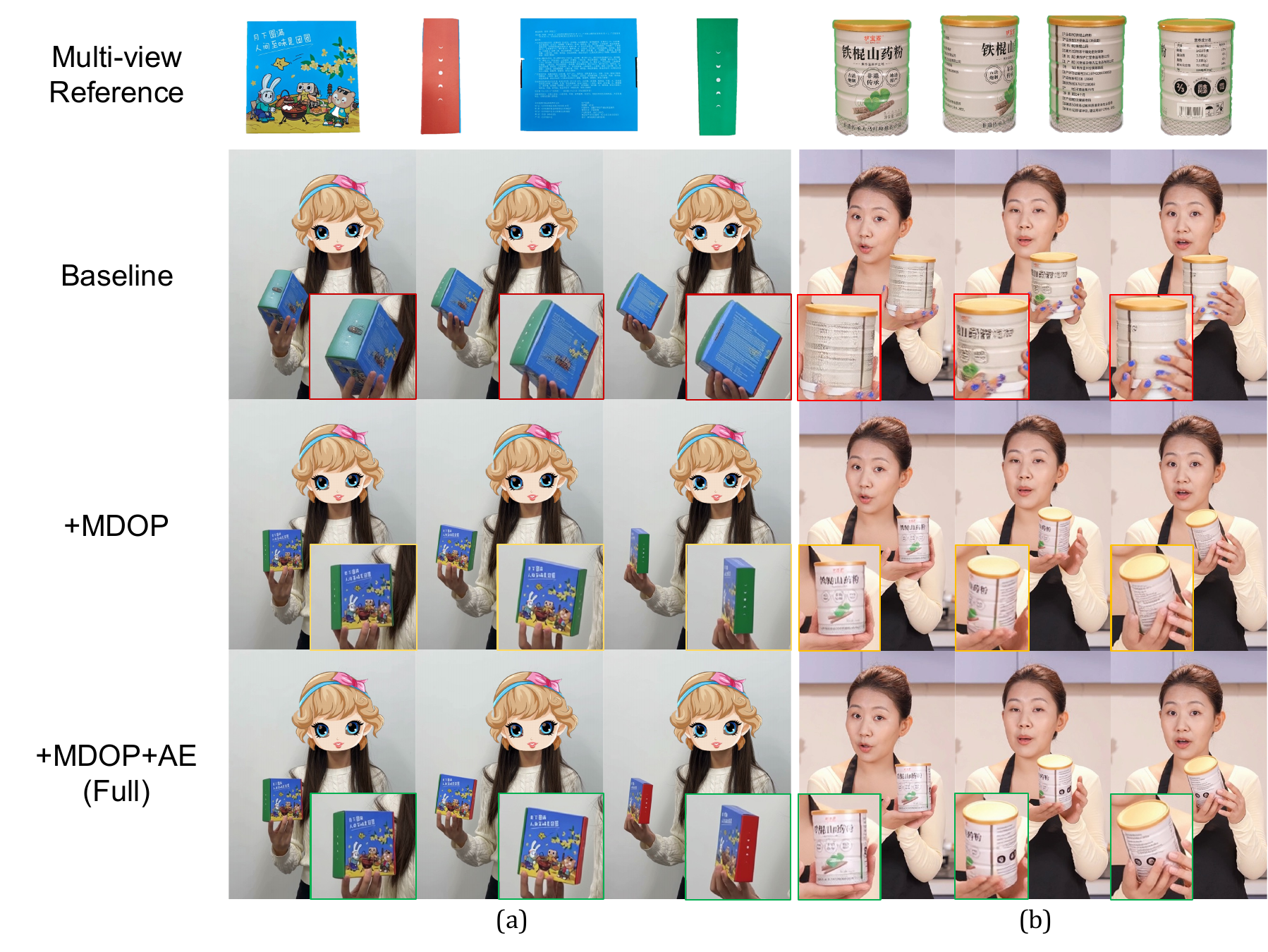}
  \caption{Qualitative comparison of the baseline, MDOP conditioning, and the full model with attention enhancement (AE).}
  \label{fig:exp_ablation}
\end{figure}

\begin{table*}[t]
  \centering
  \begin{minipage}[t]{0.47\textwidth}
    \centering
    \small
    \begin{tabular}{lccc}
      \toprule
      Method        & FID $\downarrow$ & FVD  $\downarrow$   & O-CLIP $\uparrow$ \\
      \midrule
      MimicMotion   & 87.25  & 892.2  & 0.421 \\
      VACE          & 79.23  & 618.2  & 0.592 \\
      HuMo          & 277.20 & 2530.0 & 0.333 \\
      HunyuanCustom & 84.56  & 699.0  & 0.608 \\
      GenHOI        & 72.15  & 578.2  & 0.613 \\
      Ours & \textbf{66.44} & \textbf{534.1} & \textbf{0.617} \\
      \bottomrule
    \end{tabular}
    \captionof{table}{Automatic-metric comparison on the AnchorCrafter benchmark. FID/FVD measure perceptual realism and temporal coherence (lower is better), while O-CLIP evaluates object appearance consistency (higher is better).}
    \label{tab:anchorcrafter}
  \end{minipage}\hfill
  \begin{minipage}[t]{0.47\textwidth}
    \centering
    \small
    \begin{tabular}{l|ccc}
      \toprule
      Method & FID $\downarrow$ & FVD $\downarrow$ & O-CLIP $\uparrow$ \\
      \midrule
      VACE & 52.47 & 279.1 & 0.546 \\
      HunyuanCustom & 61.48 & 364.4 & 0.532 \\
      GenHOI & 48.08 & 281.6 & 0.549 \\
      Ours & \textbf{38.15} & \textbf{267.0} & \textbf{0.557} \\
      \bottomrule
    \end{tabular}
    \captionof{table}{Quantitative comparison with baseline methods on long-video generation. The corresponding human evaluation is reported in the user-study table of the main paper.}
    \label{tab:exp_longvideo}
  \end{minipage}
\end{table*}

\Cref{fig:exp_ablation} complements the quantitative ablation in the main paper with two representative interactions. Within each example, the source interaction and multi-view target references are held fixed across the three rows, isolating the effect of the added conditions. The baseline conditions the video generator only on the multi-view target references. Although these references provide object appearance, they do not specify the frame-dependent object state; consequently, the generated object may follow an incorrect orientation, change shape across frames, or become misaligned with the driving interaction.

In the shown examples, adding the coarse anchors produced by MDOP supplies an explicit frame-dependent structural condition. The middle row follows the source manipulation more closely and preserves a more stable silhouette and object scale under rotation. However, the coarse anchors do not retain all high-frequency appearance details, so the video generator can still confuse reference viewpoints.

The full model additionally applies AE to favor references compatible with the current object state. As shown in the bottom row, it better preserves view-dependent colors, labels, and surface details while retaining the motion and geometry established by MDOP. Together, these qualitative results distinguish the two components: MDOP principally establishes the coarse motion-aligned object state, whereas AE refines the selection of appearance information from the target views. The progression is consistent with the quantitative ablation in the main paper.

\subsection{Results on the AnchorCrafter Benchmark}
\label{app:anchorcrafter}

As reported in \cref{tab:anchorcrafter}, MVHOI obtains the best value on each of the three automatic metrics. The consistent gains in distributional video quality, temporal coherence, and target-object fidelity indicate that the benefits of motion-aligned structural guidance and multi-view appearance conditioning extend to the public AnchorCrafter benchmark rather than only to the collected data.

\subsection{Long-Video Generation}
\label{app:long_video}

\paragraph{Comparison with baselines.}
We evaluate the complete framework on 10-second Cross-Reenactment sequences using matched source videos and target-product conditions. As reported in \cref{tab:exp_longvideo}, MVHOI achieves the best performance across all three automatic metrics. The consistent gains indicate that MVHOI retains both target-object appearance and video-level coherence as the generation horizon increases.

This comparison evaluates the complete MVHOI system rather than isolating the contribution of cross-iterative inference. In particular, each segment is generated from MDOP anchors and multi-view references, and the refined object state at one segment boundary initializes the next MDOP rollout. The result therefore reflects the combined effect of motion-aware prior construction, reference conditioning, and cross-iterative state refresh.

\paragraph{Effect of cross-iterative inference.}
\begin{figure}[t]
  \centering
  \includegraphics[width=0.95\linewidth]{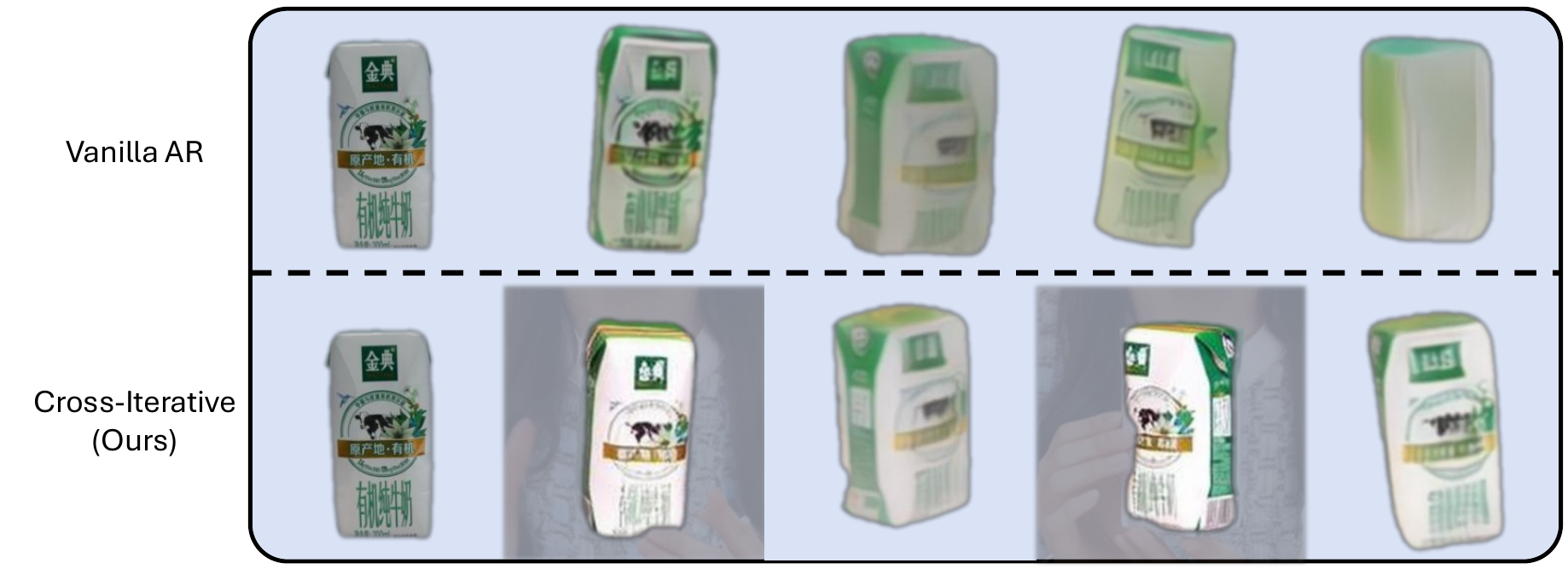}
  \caption{Qualitative comparison between vanilla autoregressive (AR) inference and cross-iterative inference, shown at an interval of 80 frames. Frames with human context in the bottom row denote refined outputs used at segment boundaries.}
  \label{fig:AR}
\end{figure}
As illustrated in \Cref{fig:AR}, vanilla AR gradually exhibits geometric distortion and severe object deformation, whereas cross-iterative inference better preserves the object's shape and texture over the same horizon. The refined frames used at segment boundaries provide subsequent MDOP rollouts with a higher-quality visual state than a purely autoregressive continuation, reducing the accumulation of coarse-anchor errors over successive clips.

\section{Limitations}
\label{app:limitations}
\paragraph{Viewpoint coverage.} The synthesis quality depends on the comprehensiveness of the multi-view references. If an object rotates to a viewpoint not covered by the reference set, the model may produce blurred textures or inconsistent appearances due to insufficient visual cues.
\paragraph{Motion extraction under occlusion.} Extreme or prolonged hand--object occlusions in the source video can hinder the motion extractor. This may lead to inaccurate 3D trajectory distillation, resulting in misaligned poses or physically implausible interactions.

\paragraph{Dependence on object masks.} The current pipeline requires object masks to extract source-object crops and to condition the masked video-generation process. Inaccurate segmentation or tracking can contaminate the motion signal, leave residual source-object content, or exclude regions needed for the replacement object. Integrating robust automatic mask estimation and uncertainty-aware conditioning would make the system more practical for unconstrained videos.


\end{document}